# Benchmarking terminology building capabilities of ChatGPT on an English-Russian Fashion Corpus

**Anastasiia Bezobrazova**

Centre for Translation Studies

University of Surrey, UK

bezobrazovaanastasia@gmail.com

**Miriam Seghiri**

IUITILM

University of Malaga, Spain

seghiri@uma.es

**Constantin Orasan**

Centre for Translation Studies

University of Surrey, UK

c.orasan@surrey.ac.uk

## Abstract

This paper compares the accuracy of the terms extracted using SketchEngine, TBXTools and ChatGPT. In addition, it evaluates the quality of the definitions produced by ChatGPT for these terms. The research is carried out on a comparable corpus of fashion magazines written in English and Russian collected from the web. A gold standard for the fashion terminology was also developed by identifying web pages that can be harvested automatically and contain definitions of terms from the fashion domain in English and Russian. This gold standard was used to evaluate the quality of the extracted terms and of the definitions produced. Our evaluation shows that TBXTools and SketchEngine, while capable of high recall, suffer from reduced precision as the number of terms increases, which affects their overall performance. Conversely, ChatGPT demonstrates superior performance, maintaining or improving precision as more terms are considered. Analysis of the definitions produced by ChatGPT for 60 commonly used terms in English and Russian shows that ChatGPT maintains a reasonable level of accuracy and fidelity across languages, but sometimes the definitions in both languages miss crucial specifics and include unnecessary deviations. Our research reveals that no single tool excels universally; each has strengths suited to particular aspects of terminology extraction and application.

## 1   Introduction

The rise of digital communication and the increasing globalisation of industries have underscored the necessity for reliable multilingual dictionaries that accurately reflect the nuances and specific terminology of various specialised fields. Specialised fields such as law, engineering, medicine, or fashion have their own terminologies and vocabularies, requiring precise and comprehensive bilingual lexicons for efficient language processing (Chodkiewicz et al. 2002). Due to the specialist character of this terminology, extracting these lexicons from specialised corpora may pose particular difficulties. Tools like SketchEngine (Kigarriff et al,

2014) and TBXTools (Oliver and Vazquez, 2015) are commonly used to extract terms from domain specific corpora, but they cannot provide definitions for the extracted terms. The recent developments in the field of Generative AI (GenAI) have attracted the attention of terminologists who proposed ways to used Large Language Models (LLMs), and ChatGPT in particular, to support the process of building terminologies (Giguere et al, 2023; Massion, 2024). In contrast to the commonly used tools for terminology extraction, GenAI can extract terms from domain-specific corpora and propose definitions for them. This characteristic was successfully employed by Lew (2023) to generate definitions for dictionary entries.

The aim of this research is to demonstrate the feasibility of compiling a reliable and high-quality corpus of fashion texts in English and Russian which can serve as a valuable source for creating a bilingual glossary that can aid translators in their work. In addition, we compare the accuracy of the terms extracted using SketchEngine, TBXTools and ChatGPT, discussing their strengths and weaknesses. Given the ability of ChatGPT to generate texts on the basis of prompts that it receives, we also evaluate the quality of the definitions produced by ChatGPT. The research presented in this paper focuses on the English-Russian language pair, but the proposed methodology can be easily adapted to other language pairs and applied to other domains.

Whilst ChatGPT has proved its usefulness in numerous applications, there is limited research that systematically assesses its ability to extract terminologies. Apart from Giguere et al. (2023) who compared the terms extracted by GPT4 with a statistical model, to the best of our knowledge, there are no other academic studies that assesses the performance of ChatGPT for terminology extraction. The majority of publications on this topic are focused on how ChatGPT can support translation professionals by providing practical guidance on the use of ChatGPT for this purpose. Good examples for this are (Muegge, 2023) and numerous posts on social media[1] and company blogs[2]. Whilst such publications can offer invaluable practical information on how to use ChatGPT for building terminologies, they do not provide comprehensive evaluations of its performance. This research aims to address this gap by carefully evaluating the performance of SketchEngine, TBXTools and ChatGPT on our corpus.

The rest of the paper is structured as follows: Section 2 presents the process of preparing the data used in this study. The three tools employed in this study are briefly described in Section 3, followed by their evaluation in Section 4. The paper finishes with a discussion and conclusions.

## 2    Data preparation

In this section we present the corpus compilation process and how the gold standards used in this research were created.

---

[1] https://www.linkedin.com/pulse/large-language-models-terminology-extraction-yannis-evangelou/
[2] https://www.oneword.de/en/term-extraction-ai/

## 2.1    Corpus compilation

The essence of the corpus-driven terminology extraction methods lies in their ability to highlight relevant terms for the domain represented in the corpus. In the case of this research, this means that provided that the corpus represents the fashion domain accurately, the terminology extracted from the corpus captures the language used within fashion-related discourses in Russian and English, (Afzaal et al. 2023). Achieving this representativeness depends very much on how the corpus used in the extraction process was compiled. Our research was carried out on a comparable corpus of magazines and webpages related to fashion written in English and Russian. This section presents the process of corpus compilation and details about the corpus.

The **first step** in corpus building entailed identifying sources of information and locating texts to be included in the corpus. The Google search engine was utilised to identify popular and reputable fashion magazines and websites renowned for their significant impact and readership. This process was carried out by the first author of the paper, who possesses expertise in the fashion domain. By analysing various resources such as '*Top 14 Fashion Magazines In The World*', '*Top 10 Luxury Fashion Magazines*', '*Fashion Magazines: History of the Biggest Magazines - Vogue, ELLE & Co.*', we determined that *Vogue* was the most important magazine globally. Subsequently, other prominent magazines like *Cosmopolitan*, *ELLE*, and *Glamour* were also selected as a source of the texts to be included in the corpus. This ensured a comprehensive and authoritative representation of the fashion domain.

In addition to these magazine texts, the corpus was enriched by the inclusion of content from 52 leading fashion websites. These websites were selected using the same criteria as the magazines, ensuring that they meet high standards of relevance, authority, and influence in the fashion domain. This expansion served to diversify the corpus's virtual nature further, by embracing the dynamic and evolving landscape of online fashion discourse. All the magazines and webpages downloaded were published between 2021 and 2024, thus ensuring that the corpus represents the language currently used in the fashion industry.

The **second step** involved downloading the chosen magazines from their web pages for future storage. This task was accomplished by right-clicking on the desired content and selecting 'Download' in PDF format. Given that our analysis also required the magazines in plain text format (TXT) for the terminology extraction programs, an online converter was employed to convert the downloaded files to the required format. In parallel to the acquisition of magazine content, digital materials sourced from the selected fashion websites were stored in HTML format. This decision was informed by the need to preserve the original formatting and interactive elements inherent to web-based content, which might be useful in future analysis of digital fashion discourse. In addition, the HTML files were converted to text format by saving them in TXT format.

The result of this process was a comparable, virtual, bilingual corpus comprising 24 fashion magazines, with 12 in Russian and 12 in English. In addition, we collected web pages from 52 websites, equally distributed between English and Russian. This composition ensures that the corpus is not only diverse in terms of content source, but also balanced linguistically, making it qualitatively representative of the fashion domain, and providing a comprehensive overview of the industry's discourse. After cleaning the corpus (see Section 2.2) we tested the representativeness of the corpus using the ReCor tool[3]. ReCor relies on lexical density to determine the minimum number of texts and words that should be included in a specialised language corpus in order to be representative for that particular domain (Corpas Pastor and Seghiri 2007). ReCor applied to our corpora confirmed that "the corpora are nearing a state of representativeness".

## 2.2    Corpus cleaning

The process of creating a corpus from fashion magazines involved downloading the PDF and HTML files and converting them to TXT format. However, the conversion process proved to be a challenging task. Despite testing numerous programs, the conversion often resulted in TXT files that were not presentable due to a plethora of issues, including the presence of symbols, numbers within words, significant gaps between words, and a slew of unnecessary symbols. This was particularly true when we converted the PDF files, due to the fact that they contained many images and employed a very creative layout for pages. The appendix contains two examples of these noisy texts.

Despite experimenting with several programs such as Text Cleaner[4], Text Tools[5], ReText.AI[6], and Code Beautify[7], the outcomes were unsatisfactory. Although these programs boasted user-friendly interfaces, they failed to deliver acceptable results. The resulting TXT files still contained significant levels of noise, or an important part of information was lost. Consequently, the decision was made to explore ChatGPT as an alternative text cleaning tool. To this end, we experimented with a number of prompts where ChatGPT was asked to remove the noise and make the text readable. Given that the corpus was collected from the Internet it is possible that the texts have been ingested by ChatGPT during its training, which made the task easier. Given the tendency of Large Language Models (LLMs) to hallucinate, the output was carefully checked to ensure that it did not contain information which was not present in the original or that no essential information was removed. Even with this manual checking step, the cleaning of the corpus was much faster with the help of ChatGPT, than if it had been done manually.

ChatGPT produced markedly better results than the other tools used, albeit with its own set of challenges. The most effective prompts for cleaning English texts included: "*Clean this text and make it readable.*" and "*Make this text readable, keep it as original as possible, remove the noise, keep all information.*"[8] For Russian texts, the approach was similar, with some prompts in English but annotated with '*in Russian*' for instance, i.e. '*make it readable in Russian*.' Without such a note, the cleaned Russian texts were at times inadvertently translated

---

into English. Additionally, the same prompt could yield different results, and occasionally, ChatGPT would unexpectedly cease providing output with the message, '*I'm sorry, but I can't assist with that request*' despite having functioned moments before. In such situations, the process had to be restarted using a new chat.

One of the challenges encountered during the cleaning process was that at times, prompts intended to clean the text and reduce noise sometimes resulted in a summary rather than the original text, though the prompt '*clean the noise, make this text readable and keep it as original as possible*' generally produced good results. Nevertheless, repeated prompting was occasionally necessary to achieve the desired outcome, and even then, the results were sometimes unsatisfactory. Switching to a new chat and repeating the exact same prompt would, however, complete the request as needed.

We also experimented with more complicated prompts that considered the characteristics of a particular source. Whilst the prompt usually worked better for that particular source, it was rarely very useful for other sources that had a completely different layout. As a result, we decided to use these generic prompts that worked across all the sources, and manually check the results. All the experiments presented in this paper were carried out using the web interface of ChatGPT. In future experiments, we plan use the API which provides more flexibility and would allow us to apply a cascade of prompts to clean a text step by step. This may allow us to automate the process more.

For cleaning, we experimented with both ChatGPT-3.5 and ChatGPT-4. In the initial stages of this research, only ChatGPT-3.5 was available, but we switched to ChatGPT-4 as soon as it became available. The majority of text cleaning was performed with ChatGPT-4. Overall, the text cleaning process was time-consuming as it had to be done manually, one text at a time, but this approach was the only one that provided satisfactory results. The resulting corpus contains over 1.8 million words, with 1 million words in English and 800,000 words in Russian.

## 2.3   Gold standard development

To effectively evaluate the terms extracted by term extraction tools such as SketchEngine, ChatGPT, and TBXTools, it was necessary to establish a robust gold standard.[9] For this reason, we used the Google Search engine to locate pages containing lists of fashion related terms and harvested them using bespoke python scripts. The selected pages were relatively easy to harvest as the terms were listed using clearly structured tables. At times these tables also contained the translation of terms and their definitions.

The gold standard developed comprises a carefully curated list of terms pertinent to the fashion domain. The terminology was primarily sourced through an automated harvesting process from bilingual English-Russian websites, alongside monolingual English sites. The terms harvested from English only websites were automatically translated and rigorously corrected by a native speaker to ensure their accuracy and relevance within both linguistic frameworks.

---

[9] The term *gold standard* is used in the field of Natural Language Processing to refer to a resource that was validated by humans and is used to evaluate automatic processing methods by comparing their output with the gold standard using establish comparison methods.

The gold standard contains a total of 354 terms in English and Russian. Sixty of these terms also had definitions in English. As with the terms, we automatically translated the definitions to Russian and carefully checked their accuracy. The definitions were used to assess the ChatGPT's ability to extract definitions.

## 3    Terminology extraction tools

Once the data was preprocessed, terms were identified in the corpora with the help of SketchEngine, TBXTools, and ChatGPT. These tools use various algorithms and techniques to identify frequent and domain-specific terms, facilitating a comprehensive analysis of the terminology within the fashion industry.

SketchEngine is a widely used tool for terminology extraction in academic circles, owing to its highly customisable features and comprehensive linguistic resources encompassing corpora, dictionaries, and thesauri. Its advanced querying capabilities facilitate precise and efficient extraction, utilising linguistic templates and domain-specific terms. Moreover, the tool incorporates phrase analysis functionality to identify commonly occurring phrases, while its evaluation and validation tools guarantees the quality and reliability of the extracted terms by conducting comparisons with external resources or expert knowledge (Kilgarriff et al. 2014).

TBXTools is a highly capable tool for terminology extraction, offering a range of functionalities. It employs statistical and linguistic methods to extract multiword terms from specialised corpora. The tool supports statistical term extraction using n-grams and stop words, linguistic term extraction using morpho-syntactic patterns and a tagged corpus, detection of translation candidates in parallel corpora, and automatic learning of morphological patterns. Additionally, it offers evaluation capabilities to assess precision and recall based on different frequency thresholds. TBXTools proved its effectiveness in various language processing tasks, including ontology learning, machine translation, computer-assisted translation, thesaurus construction, classification, indexing, information retrieval, and text mining. Its Python-based nature enhances usability, flexibility, and compatibility across platforms and systems, making it a valuable asset for researchers and practitioners in the field of terminology management. The fact that it is open source enables other researchers to extend it and adapt it as needed (Oliver and Vàzquez 2015).

ChatGPT is not specifically designed for terminology extraction, but it can extract terms if prompted correctly. We experimented with a variety of prompts such as:

> *Extract the terms related to fashion, such as all kinds of clothes, shoes, accessories etc from the given text and list them. Can you please extract terms, that can be found ONLY in the given text*[10]

---

[10] https://chat.openai.com/share/4e30b4e7-1086-4df2-ac66-8b703e4ae17d

One limitation of ChatGPT is that it demands considerable time to extract an extensive array of terms. Initially, ChatGPT may provide between 20 to 50 highly relevant terms. To elicit further terminology, it is necessary to continue prompting iteratively. The process was repeated till ChatGPT started giving repetitive words, non-existing words, colour plus clothing items, or unrelated words like "shower". During the process it was noticed that ChatGPT can sometimes deviate from the specific corpus under consideration, beginning to extract domain-specific terminology—such as fashion terms—in a more general context without relying on the text provided.

## 4 Results

### 4.1 Terminology extraction

In this section we compare the performance of the SketchEngine, TBXTools and ChatGPT at extracting terms from our corpus. The performance is calculated using standard metrics like precision, recall, and f-measure. **Precision** measures the accuracy of the extracted terms by evaluating how many of the extracted terms are in our gold standard. A high precision indicates that fewer irrelevant terms are extracted. **Recall** measures the completeness of the term extraction by assessing how many of the terms from the gold standard were extracted. High recall means that most of the terms from the gold standard were extracted. **F-measure** is the harmonic mean of precision and recall, providing a single metric to balance the two. The results for these metrics for each language are presented in Table 1. The analysis presented in this section will highlight the strengths and weaknesses of each tool, with a particular emphasis on how well they balance precision and recall in the extraction process.

| | English corpus | | | Russian corpus | | |
|---|---|---|---|---|---|---|
| | Precision | Recall | F-measure | Precision | Recall | F-measure |
| **TBXTools** | 0.022 | 0.704 | 0.044 | 0.055 | **0.833** | 0.104 |
| **SketchEngine** | 0.009 | **0.732** | 0.018 | 0.008 | 0.629 | 0.016 |
| **ChatGPT** | **0.283** | 0.360 | **0.317** | **0.335** | 0.358 | **0.346** |

Table 1: The accuracy of term extraction using the three tools and measured using precision, recall and f-measure. The values in bold represent the highest values observed for a metric and a corpus

TBXTools extracted approximately 15,000 terms in English and 7,800 terms in Russian, with frequencies ranging from a minimum of 1 to a maximum of 4. After we applied an automatic cleaner, which removed superfluous characters such as dashes, 'at', and numerals, standardised plurals and converted the terms to lowercase the total number of terms was reduced to approximately 10,000 in English and 5,000 in Russian. Table 1 presents the accuracy of TBXTools. The results from the TBX Tools term extraction exhibit a high recall, especially for Russian, suggesting the tool's efficacy at identifying a broad array of terms, including a large proportion of relevant terms from the gold standard.

SketchEngine extracted approximately two to three times as many terms in English and Russian compared to TBXTools, resulting in a higher incidence of noise. In the case of English, the predominant issue was the amalgamation of separate phrases, exemplified by terms such as 'andjeans.' For Russian, the errors were twofold: some terms appeared in English rather than Russian, and others where single words were incorrectly split into two words. Following both automatic and manual cleaning processes, the number of terms was reduced to approximately 10,000 for each language. Even after cleaning the lists, the precision remains low for both languages, as can be seen in Table 1. SketchEngine shows a notable ability to achieve high recall in both English and Russian, effectively capturing a large proportion of relevant terms. This is not surprising given the large number of terms it extracts. However, it struggles significantly with precision, with the inclusion of many irrelevant terms leading to a high incidence of noise.

As mentioned above, one of the challenges of using ChatGPT for extracting terms was that it had to be prompted repeatedly in order to produce a longer list of terms as each time the list contained between 20 to 50 terms. To elicit further terminology, it is necessary to continue prompting iteratively. Additionally, ChatGPT can sometimes deviate from the specific corpus under consideration, beginning to extract domain-specific terminology - such as fashion terms - in a more general context without relying on the provided text. No cleaning was applied to the list of terms produced by ChatGPT. As can be seen in Table 1, ChatGPT obtains a significantly higher precision and f-measure scores, but the recall is the lowest. This can be explained by the fact that the number of terms extracted using ChatGPT was in the hundreds, rather than thousands as TBXTools and SketchEngine produce.

A notable advantage of using ChatGPT is its ability to classify terms into distinct categories. We noticed that at times ChatGPT would organise the extracted terms into categories such as 'clothes', 'shoes', 'accessories', and 'bags'. This capability facilitates a more structured approach to understanding and organising terminology, which can be particularly beneficial for academic and research purposes in specialised fields.

## 4.2 Evaluation of ranking of terms

TBXTools and SketchEngine return lists of terms ranked according to their term likelihood. Because our gold standard contains only 354 terms, we decided to evaluate how increasing the number of terms we consider up to 354 terms impacts on the accuracy of the terms extracted. This would simulate a scenario where a terminologist uses automatic tools to build a glossary, and they consider the extracted terms in the order returned by the tools.

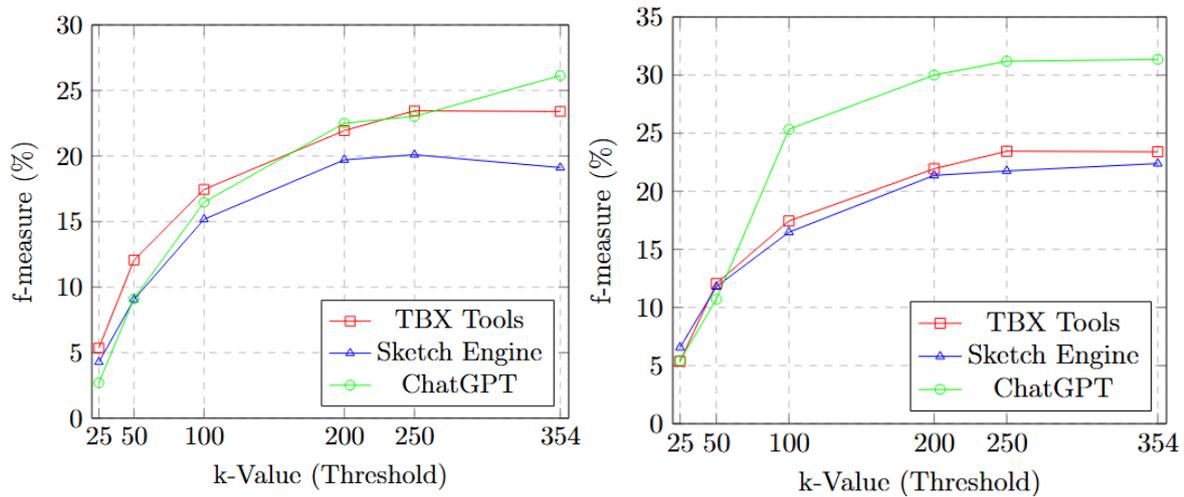

Figure 1: Changes in the f-measure scores as we increase the number of terms considered for the English corpus (left) and Russian corpus (right)

Figure 1 demonstrates how the scores for f-measure change as we increase the number of terms we consider up to the size of the gold standard. Due to space reasons, we do not include the graphs presenting the precision and recall scores. While analysing these scores, we noticed that in most of the cases as the number of terms we consider increases, the recall also increases at the expense of precision. The exception is when we extract terms using ChatGPT. As we increase the number of terms from 25 to 100, both precision and recall increase (precision from 0.20 to 0.40, recall from 0.01 to 0.10). If the number of terms is increased further, the pattern observed with the other term extraction methods is followed i.e. the precision decreases whilst the recall keeps increasing.

Analysis of the f-measure scores across both languages shows that SketchEngine typically has lower scores, reflecting the challenge in maintaining a balance between high recall and lower precision. This is evident in both English and Russian, where the f-measure scores increase with k, but remain modest due to the precision drop. TBXTools consistently achieves slightly better f-measure scores than SketchEngine, though it is still hampered by low precision levels. The f-measure scores improve as more terms are considered, peaking at k=250 before the decline in precision affects the score adversely. ChatGPT achieves higher f-measure scores in both language datasets, indicating a better overall balance between precision and recall. This suggests that ChatGPT might be more effective when managing the extraction scope through specific prompting and adjustments, adapting better to the nuances of each language compared to the more static algorithms of TBXTools and SketchEngine.

### 4.3 Error analysis of terminology extraction

In order to understand better how good SketchEngine, TBXTools and ChatGPT are at term extraction we conducted a detailed error analysis. For this purpose, we analysed the 50 top terms extracted by each of the tools. These terms were compared against our fashion-specific gold standard to identify deviations and inaccuracies in fashion terminology. The objective being to pinpoint significant discrepancies.

Analysis of the terms extracted by TBXTools and SketchEngine reveals that they extract a large number of words that are not terms. TBXTools favours extraction of words related to the fashion industry, but which are too general to be included in our gold standard. Examples of such words are terms such as 'collection,' 'design,' 'colour,' 'мода' (fashion), 'цвет' (colour), 'одежда' (clothing). Most of the terms correctly extracted by these tools, but not included in the gold standard, are names of well-known fashion brands such as Chanel, Gucci, Prada, Versace, and Balenciaga.

The comparison between ChatGPT's fashion-related terminology and the gold standard highlights ChatGPT's effectiveness in capturing both contemporary and niche terms not included in traditional terminologies. All the top 50 terms extracted by ChatGPT are fashion-related terms in English and only one is not a term in Russian. However, a large proportion of these terms are not present in our gold standard which shows one of the limitations of employing automatic evaluation metrics.

### 4.4    Definition extraction using ChatGPT

We also analysed the definitions produced by ChatGPT for 60 commonly used terms in English and Russian. As a reference, we used the definitions we extracted from the online glossaries. The definitions for the extracted terms were produced by prompting ChatGPT to define a term in the context of the fashion industry and given our corpus.

The effectiveness of ChatGPT in providing definitions is examined by measuring the similarity between the model-generated definitions and the reference definitions using the Levenshtein distance (Levenshtein, 1966). This metric, quantifying the minimum number of edits needed to change one sequence into another, serves as an indicator of how closely the definition by ChatGPT match the expected text. We calculated the distance at the word level, rather than character level. We decided to use Levenshtein distance because it shows the number of word level edits a terminologist needs to make in order to produce the definition from the gold standard.

Significant variations in Levenshtein distances were observed, with values ranging from as low as 1 to as high as 221. The lowest distances indicate cases where ChatGPT's definitions are almost identical to the reference, demonstrating high fidelity in reproducing accepted definitions with minimal alteration. This could suggest that the pages used to extract the reference definitions are included in ChatGPT. At the other end are the highest distances which reflect instances where the definitions have been substantially modified, suggesting that ChatGPT has either added extraneous information or shifted the focus of the definition, potentially leading to deviations from the intended meanings.

For the English definitions, the lowest recorded distance is 1, where ChatGPT made a minor modification by replacing 'usually' with 'typically', showing high fidelity in reproducing the reference definition almost identically. In contrast, the highest distance observed is 221, where ChatGPT expanded significantly on the reference by adding various contextual details, leading to a potential shift away from the original meaning and introducing possible inaccuracies. The average Levenshtein distance between the reference descriptions and definition by ChatGPT is 14.91 indicating that we would need an editor to make around 15 word changes per definition. For the Russian definitions, the Levenshtein distances, range from as low as 0 to as high as 94, with an average distance of 8.69 tokens.

In order to gain a better understanding of the differences between the definitions produced by ChatGPT and the ones in our gold standard, we carried out a detailed analysis and noticed the following phenomena across both languages:

**Handling of Core Concepts**: ChatGPT generally retains the core concepts of the terms it defines. For instance, regardless of language, the descriptions for items like '*jacket*', '*sweat-shirt*', and '*coat*' maintain essential elements such as their use and basic form (e.g., long sleeves, upper body coverage).

**Synonym Replacement and Structural Changes**: Changes often involve synonyms or slight structural adjustments without dramatically altering meaning, like changing '*usually*' to '*typically*' which affects the Levenshtein distance, but not the overall meaning.

**Elaboration**: ChatGPT often elaborates on the reference. For instance, the reference might simply describe an item as '*a jacket made of wind-resistant material*' while ChatGPT might expand this to '*a thin lightweight jacket designed to resist wind chill and light rain*'. This adds context and details which are not present in the reference leading to high Levenshtein distance. The English definitions tend to be more expanded and detailed compared to the Russian ones.

**Specification and Detailing**: ChatGPT descriptions tend to specify materials, contexts, or uses, such as changing '*a long thick coat worn in cold weather*' to '*a long warm coat worn over other clothing in cold weather*'. At times, this can significantly alter the semantic meaning and applicability of the descriptions.

**Omission of Specifics**: There are instances where ChatGPT omits specific details that could be crucial for a complete understanding of the term. This occurs in both languages, such as missing the '*strong blue cotton cloth*' for jeans or omitting '*military style*' from the description of a trench coat.

**Quality of Additional Information**: In both languages, the added details can either enrich the understanding of a term or potentially lead to inaccuracies. The effectiveness of these additions depends on the context in which the definition is used. In educational or technical contexts, such precision and additional context may be valuable, but it could also complicate understanding in more general uses.

In summary, while the definitions produced by ChatGPT maintain a reasonable level of accuracy and fidelity across languages, the English definitions exhibit a greater degree of elaboration and variability. ChatGPT shows proficiency in handling core concepts but could benefit from improvements in consistently including crucial specifics and managing the extent of elaboration to avoid unnecessary deviations. This analysis underscores the importance of fine-tuning and potentially adjusting the model outputs based on the target language and the specificity required by the usage context. Moreover, it is important to recognise that automatic evaluations, such as those using the Levenshtein distance, can sometimes yield lower results even when definitions are grammatically and factually correct. This discrepancy can arise because the definition might use synonyms or include specific details that do not match the reference exactly but are still accurate. Such instances highlight the need for nuanced interpretation of evaluation metrics to appreciate the full accuracy and utility of the definitions provided.

## 5    Conclusions

This paper has presented the process of building a domain specific corpus and an evaluation of the terminology extraction tools TBXTools, SketchEngine, and ChatGPT on this corpus, providing insights into the capabilities and limitations of each tool. TBXTools showed a high recall rate, capturing a broad array of relevant terms, yet struggled with precision due to the inclusion of many irrelevant terms. This was seen across various k values (i.e. the number of terms extracted) in both English and Russian, with precision diminishing as more terms were considered. SketchEngine exhibited similar characteristics to TBXTools. ChatGPT presented a distinct advantage in the extraction process due to its tailored approach to terminology extraction based on the input prompts. This resulted in higher precision, particularly evident when the terms were directly linked to the input context, reducing the inclusion of irrelevant terms significantly. The performance of ChatGPT was robust across different k values, suggesting its effectiveness in handling extensive datasets without a significant loss in term relevance or accuracy. An error analysis focusing on the top 50 terms from each tool further underscored the differences in output quality. ChatGPT's outputs were particularly notable for their relevance and contextual accuracy, reflecting the model's strength in generating pertinent content based on nuanced understanding of the domain-specific texts.

Overall, the study suggests that while traditional tools like TBXTools and Sketch Engine can efficiently capture a wide range of terms, they require improvements in precision to reduce noise. While capable of extracting numerous candidate terms, the lists produced are simply too long in many contexts and include too many words that are not real terms. This may hinder the process of terminology creation. ChatGPT, on the other hand, demonstrates a strong capacity for generating both high-quality terminology lists and accurate definitions, making it a compelling choice for stakeholders in the fashion industry looking for reliable and context-aware terminology extraction solutions. However, the lists of candidates produced by ChatGPT are significantly shorter which means they may miss important terms. This analysis highlights the importance of selecting the appropriate tool based on the specific needs and contexts of terminology extraction tasks, especially in specialised fields such as fashion where accuracy and relevance are crucial.

# Appendix

Examples of noisy output in English and Russian produced when the PDF files were converted to TXT format.

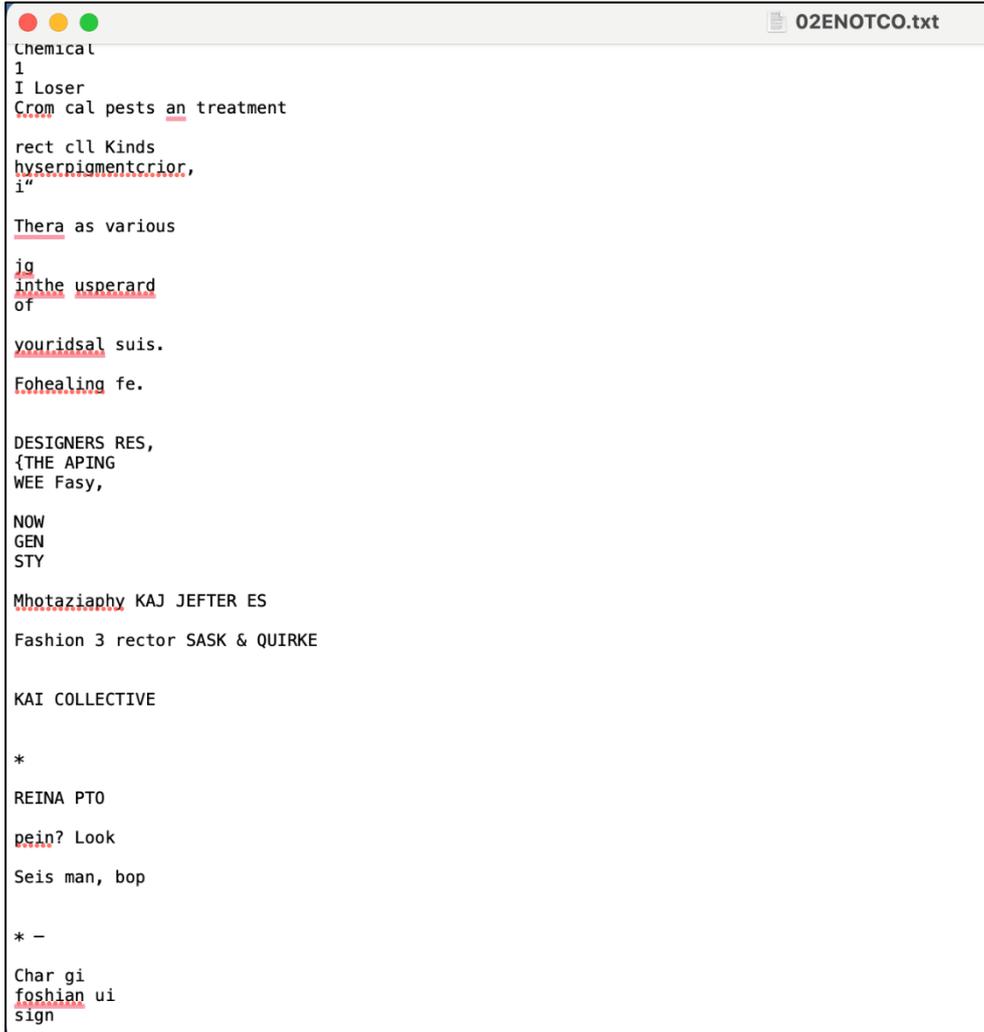

Любит: ЭКСТРИМ,
КРАСИВЫЕ МАШИНЫ
И КОРЕЙСКУЮ КУХНЮ.

УК: АМАТОШНУТ5 ОУ

ты а 1: ый '

<
————
0
ЫЧ
————
[а
<
<.
ыЭ
х.
>
о
0
0
№
[2 |
[а
=
0
–
0
®

|". "|
что слушает

Анатолий,
сканируй
ОК-код.

_й

Ме 2

Е
<
5
е

" но, Манн,
Ра — Ч